\documentclass{article}

\usepackage{PRIMEarxiv}

\usepackage[utf8]{inputenc} 
\usepackage[T1]{fontenc}    
\usepackage{hyperref}       
\usepackage{url}            
\usepackage{booktabs}       
\usepackage{amsfonts}       
\usepackage{nicefrac}       
\usepackage{microtype}      
\usepackage{lipsum}
\usepackage{fancyhdr}       
\usepackage{graphicx}       
\graphicspath{{media/}}     
\usepackage{caption}
\usepackage{subcaption}
\usepackage{amsmath}
\usepackage{natbib}

\usepackage[table]{xcolor}
\usepackage{array}
    \newcolumntype{P}[1]{>{\centering\arraybackslash}p{#1}}
    \newcommand\ChangeRT[1]{\noalign{\hrule height #1}}

\definecolor{readgreen}{RGB}{0,150,0}
\definecolor{RawSienna}{cmyk}{0,0.72,1,0.45}

\usepackage{hyperref}

\hypersetup{
    colorlinks=true,
    linkcolor=blue,
    filecolor=magenta,      
    urlcolor=blue,
    citecolor=magenta,
    pdfpagemode=FullScreen,
    }
\title{pyRDDLGym: From RDDL to Gym Environments
}

\author{%
  Ayal Taitler \\ 
  University of Toronto, CA \\
  \texttt{ataitler@gmail.com} \\
  \and
  Michael Gimelfarb \\
  University of Toronto, CA \\
  \texttt{mike.gimelfarb@mail.utoronto.ca} \\
  \and
  Jihwan Jeong \\
  University of Toronto, CA \\
  \texttt{jhjeong@mie.utoronto.ca} \\
  \and
  Sriram Gopalakrishnan \\
  J.P. Morgan AI Research \\
  \texttt{sriram.gopalakrishnan@jpmchase.com} \\
  \and
  Martin Mladenov \\
  Google, BR \\
  \texttt{mmladenov@google.com} \\
  \and
  Xiaotian Liu \\
  University of Toronto, CA \\
  \texttt{xiaotian.liu@mail.utoronto.ca} \\
  \and
  Scott Sanner \\
  University of Toronto, CA \\
  \texttt{ssanner@mie.utoronto.ca} \\
}

\begin{document}
\maketitle

\begin{abstract}
We present pyRDDLGym, a Python framework for the auto-generation of OpenAI Gym environments from RDDL declarative description. The discrete time step evolution of variables in RDDL is described by conditional probability functions, which fit naturally into the Gym step scheme. Furthermore, since RDDL is a lifted description, the modification and scaling up of environments to support multiple entities and different configurations becomes trivial rather than a tedious process prone to errors. We hope that pyRDDLGym will serve as a new wind in the reinforcement learning community by enabling easy and rapid development of benchmarks due to the unique expressive power of RDDL. By providing explicit access to the model in the RDDL description, pyRDDLGym can also facilitate research on hybrid approaches to learning from interaction while leveraging model knowledge. We present the design and built-in examples of pyRDDLGym, and the additions made to the RDDL language that were incorporated into the framework.
\end{abstract}


\section{Introduction}
Reinforcement Learning (RL) \cite{sutton2018reinforcement} and Probabilistic planning \cite{puterman2014markov} are two research branches that address stochastic problems, often under the Markov assumption for state dynamics. The planning approach requires a given model, while the learning approach improves through repeated interaction with an environment, which can be viewed as a black box. Thus, the tools and the benchmarks for these two branches have grown apart. Learning agents do not require to be able to simulate model-based transitions, and thus frameworks such as OpenAI Gym \cite{brockman2016openai} have become a standard, serving also as an interface for third-party benchmarks such as \citet{todorov2012mujoco}, \citet{bellemare2013arcade} and more.

As the model is not necessary for solving the learning problem, the environments are hard-coded in a programming language. This has several downsides; if one does wish to see the model describing the environment, it has to be reverse-engineered from the environment framework, complex problems can result in a significant development period, code bugs may make their way into the environment and finally, there is no clean way to verify the model or reuse it directly. Thus, the creation of a verified acceptable benchmark is a challenging task.

Planning agents on the other hand can interact with an environment \cite{Sanner:RDDLsim}, but in many cases simulate the model within the planning agent in order to solve the problem \cite{prost2012}.
The planning community has also come up with formal description languages for various types of problems; these include the Planning Domain Definition Language (PDDL) \cite{aeronautiques1998pddl} for classical planning problems, PDDL2.1 \cite{fox2003pddl2} for problems involving time and continuous variables, PPDDL \cite{Bryce2008InternationalPC} for classical planning problems with action probabilistic effects and rewards, and Relational Dynamic Influence Diagram Language (RDDL) \cite{Sanner:RDDL} for describing MDPs and POMDPs. While agents can use the models described by these languages to simulate transitions and compute plans, it is also natural to leverage these description languages to auto-generate environments by decoupling the mathematical problem description from the environment generation.

In recent years auto-generation tools have emerged for specific description languages, allowing for both auto-generation of environments, and access to the problem formal model, serving as a bridge between planning and learning methods to interact with a single framework. \textit{RDDLSim} \cite{Sanner:RDDLsim} is a long-standing independent framework that translates RDDL problems into an interactable environment. RDDLSim is a Java simulator, with a unique API, that requires interacting agents to manage TCP/IP connections. Although this is suitable for International Planning Competitions (IPC), the entry bar for rapid RL research is high. A more recent Python version of RDDLSim was developed to support mainly MDPs with continuous variables \cite{thiagopbueno2020}. As it had implemented the OpenAI Gym interface this tool was named \textit{rddlgym}. Following that approach, \textit{PDDLGym} \cite{silver2020pddlgym} was introduced. A Python tool that generates Gym environments from PDDL domain and problem files. Since PDDLGym works on PDDL files it can generate classical planning problems, i.e., deterministic problems. PDDLGym has some support for PPDDL which allows it to simulate action probabilistic effects, but state noise, concurrency, observations, and more components of the Markov model are not supported.

In order to be able to describe MDPs and POMDPs in a general way, allowing for factored descriptions, we present in this paper \textit{pyRDDLGym}, a Python framework for auto-generation of Gym environments from RDDL description. The library is available at \url{https://github.com/pyrddlgym-project/pyRDDLGym}. pyRDDLGym supports a Major subset of the RDDL language and in addition, has extended it to support for terminal states in MDPs. pyRDDLGym differs from rddlgym as it allows for derived-fluents, observations, discrete fluents, and more. It is also the only framework currently that supports state exogenous and endogenous noise, action concurrency, observations, and other blocks required for fully describing MDPs. We hope that pyRDDLGym will encourage more collaboration between the RL and Planning communities. This will allow hybrid methods leveraging both model description and interactions to emerge. Moreover, we aim to build a verified benchmark of stochastic domains for the probabilistic planning and RL communities.  

\begin{figure}[t]
\centering
\includegraphics[width=1.0\textwidth]{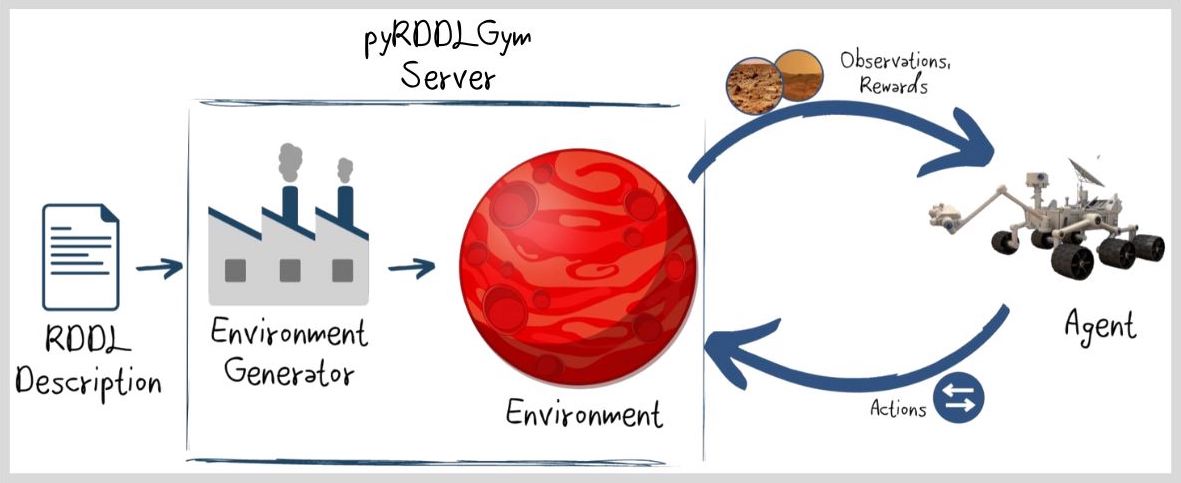}
\caption{pyRDDLGym design concept.}
\end{figure}

\section{RDDL Support and Extensions} \label{sec:subset}
RDDL \cite{Sanner:RDDL} is a \textit{lifted} declarative language to describe MDPs, where states, actions, and observations (whether discrete or continuous) are parameterized variables. RDDL leverages parameterized variables, which helps with scaling up domains. These are simple templates for ground variables that can be obtained when given a particular problem instance defining possible domain objects. The evolution of a fully or partially observed stochastic process is specified via conditional probability functions (CPFs) over the next state variables conditioned on the current state and action variables, with allowed concurrency. The objective function in RDDL is defined by the immediate rewards and the discount factor one specifies. For a grounded model (instance), RDDL is just a factored MDP, or POMDP, if partially observed.

Thus, a \textit{grounded} RDDL problem can be fitted into the Gym scheme of interaction where an agent acts and receives observation and reward from the environment. In this black box scenario, the explicit structure of the problem is lost and left for the agent to reason about. However, agents that can leverage the information in the model have the potential to boost their performance significantly.

Like all languages, RDDL also evolves in order to avoid ambiguity and increase expressibility. Also, in order to fit a Gym scheme additional features have been introduced. The deviation from the original language description \cite{Sanner:RDDL} is listed in the next section.

\begin{figure} [t]
\centering
\includegraphics[width=1.0\textwidth]{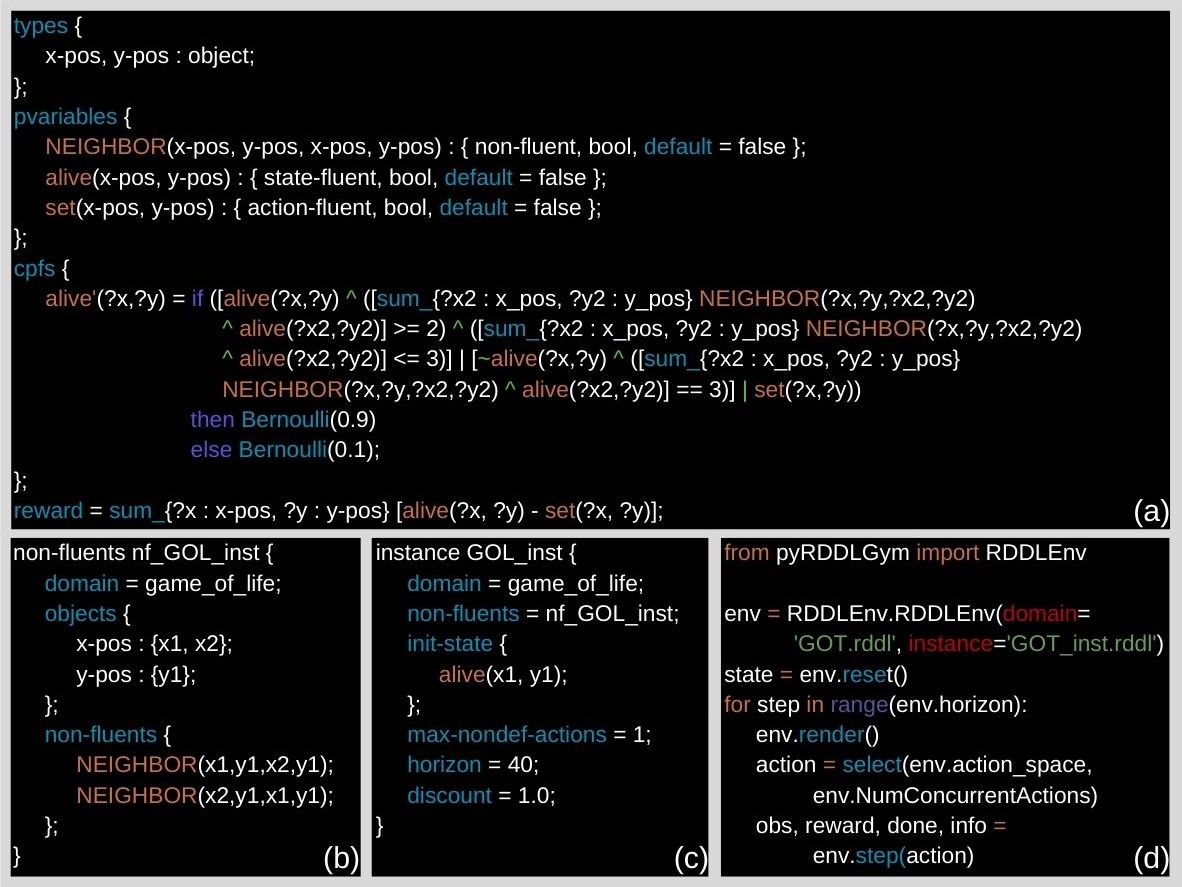}
\caption{pyRDDLGym code examples. A pyRDDLGym environment is characterized by an RDDL domain file and an instance \& non-fluents file. (a) Fluents, CPFs, and reward for the game of life problem (b) A non-fluents block, defining a two neighboring cell topology. (c) An instance block defining the parameters and init-state of the problem. (d) Using the domain and instance files, an RDDLEnv can be initialized. Interaction is similar to that of any OpenAI Gym environment.}
\end{figure}

\subsection{Language Variant}
pyRDDLGym supports the majority of the original RDDL. The following components are omitted (or marked as deprecated) from the language variant implemented in pyRDDLGym:
\begin{itemize}
    \item derived-fluent are supported by the framework as described in the language description. However, they are considered deprecated and will be removed from future versions.
    
    \item fluent levels are deprecated and are reasoned automatically by the framework as described in the next section.
    
    \item state-action-constraints are not implemented and considered deprecated in the language to avoid ambiguity. Only the newer syntax of specifying state-invariants and action-preconditions is supported.
\end{itemize}

Additional components and structures have been added to the language to increase expressivity, and to accommodate learning interaction type. These are listed here:

\begin{itemize}
\item Terminal state description has been added, described in details in the following sections.

\item action-preconditions are implemented according to \citet{Sanner:RDDL}. However, they are subject to user preference. By default the framework \textit{does not} enforce the expressions in the action-preconditions block. Thus, upon violation, a warning will be printed to the user and the simulation will push the actions inside the legal space by using the default value and the simulation will continue. To ensure correct behavior it is expected from the domain designer to include the appropriate logic of how to handle an invalid action within the \textit{CPFs} block. In the case where the user does choose to enforce action-preconditions, the simulation will be interrupted and an appropriate exception will be thrown.

\item Direct inquiry of variable (states/action) domains is supported through the standard action\_space and state\_space properties of the environment. For this functionality to work correctly, the domain designer is required to specify each (lifted-)variable bound within the action-preconditions block in the format "\textit{fluent} OP BOUND" where $\mbox{OP} \in \{<,>,>=,<=\}$, and BOUND is a deterministic function of the problem parameter to be evaluated at instantiation.

\item Parameter inequality is supported for lifted types. I.e., the following expression $?p == ?r$ can be evaluated to True or False.

\item Nested indexing is now supported, e.g., $fluent'(?p,?q) = NEXT(fluent(?p, ?q))$.

\item Vectorized distributions such as Multivariate normal, Student, Dirichlet, and Multinomial are now supported.

\item Basic matrix algebra such as determinant and inverse operation are supported for two appropriate fluents.

\item $argmax$ and $argmin$ are supported over enumerated types (enums).

\end{itemize}

\subsection{Level Reasoning} \label{sec:levels}
Formally in RDDL when a derived or an interm fluent is declared, its level should also be stated to define its position in the Dynamic Bayes Net (DBN) of the problem, which dictates the fluents' order of evaluation. While the level hierarchy declaration is not overly complicated from the domain designer's side, it is completely unnecessary. In pyRDDLGym, following the implementation of \citet{Sanner:RDDLsim}, this declaration is omitted, and ignored if supplied. 
The order of fluent evaluation at any given time step is always as follows
\begin{equation} \label{eq:order}
    s_t \rightarrow d_t \rightarrow i_t \rightarrow s_{t+1},
\end{equation}
where $s_t$ denotes the current state, $d_t$ denotes the derived fluents, $i_t$ denotes the interm fluents, and $s_{t+1}$ denotes the next state. A requirement when designing a domain is that the fluent order must form a directed acyclic graph (DAG). Thus, we first generate a call graph from the CPFs, for all fluent types. Then we use topological sorting to sort the fluents by the order of evaluation. This has two merits, the first is the reasoning over the over of fluent evaluation, and second, validation that there are no cycles in the evaluation order, and that the correct order of evaluation in \eqref{eq:order} is conserved.

\subsection{Terminal States} \label{sec:termination}
Another extension to RDDL is the addition of terminal states to the language. An MDP may have a terminal state, which can be in the form of a goal state or a state where there are no available actions anymore. The key is that in both of these cases, it is desired to end the simulation, e.g., an agent hits a wall, a pendulum falls beyond a threshold, etc. RDDL currently does not support these cases, only fixed horizon problems. On the other hand, Gym can terminate an episode anytime with the \textit{done} flag.

Thus, an additional block has been introduced into RDDL to support terminal states. The keyword denoting the terminal states block is \textit{termination}, and it supports a list of conditions:
\begin{equation}
    termination\ \{ cond_1;\ cond_1;\ ...\ cond_N \};
\end{equation}
where the $cond_i,\ i\in\{1,..,N\}$ are Boolean conditions, and the full termination condition is a disjunction over all the inner conditions, i.e.,
\begin{equation}
    termination = cond_1 \lor cond_2 \lor ... \lor cond_N.
\end{equation}

\section{Design and Implementation}
In order to be compatible with the Gym API, five methods need to be implemented. First, \textit{\_\_init\_\_ ()}, which initializes the environment, and in that method, all the parsing of the RDDL, grounding, and fitting into the environment scheme is done. 
\textit{reset()} in which the environment is returned to the initial state, and the initial state is returned or \textit{None} in case of a POMDP. \textit{step()} in which the transition function is calculated according to the CPFs in the RDDL file.
\textit{render()} in which the visualization is implemented, and finally the \textit{close()} method in which resources are being freed so the simulation can be terminated. 
Two more constructs to be implemented are the properties \textit{action\_space} and \textit{observation\_space}, which inform the agent about the type and domain of the action and observation spaces respectively.

\subsection{Instantiation of Gym Environments}
The RDDL description contains three components. The first is the \textit{domain} block, in which information about the lifted domain is provided, e.g., types, CPFs, fluent definitions, etc. The second component is the \textit{instance} block which specifies a specific problem, i.e., all that is needed in order to ground the lifted abstract domain. The third block is the non-fluents block, which is being pointed at from the instance block as it is also part of how to instantiate a specific problem. This block is the constant of the instance, or more precisely it defines the topology of the problem. To create an environment all three are required. Thus, the domain containing the global problem definitions should be provided in a ".rddl" file, and the other two blocks, as they together define an instance, should be placed in a second separate ".rddl" file. The \textit{\_\_init\_\_()} method requires both these files in order to generate a Gym environment.  


The call to the \textit{\_\_init\_\_()} method, invokes first the parsing of the RDDL description provided in the domain, instance, and non-fluents, then the grounder is invoked on the parsing tree in order to generate a specific instance as required. At this point, the transformation from RDDL types to Gym spaces is done and the action and observation properties are populated. The CPF sampler object is also instantiated here for the grounded problem. In pyRDDLGym three additional properties are provided. The first is \textit{horizon}, which informs the agent about the horizon of the problem, assuming no terminal state is encountered. The value returned by this property is the number of horizon steps as defined in the RDDL horizon field in the instance block. The second property is \textit{NumConcurrentActions} which denotes the maximum number of concurrent actions the agent can send to the framework in a single time step. In the case where the field \textit{max-nondef-actions} in the RDDL instance specify \textit{pos-inf}, the number of all available grounded actions in the problem is returned, indicating that there is no limit on the number of concurrent actions. The third and last property is \textit{discount}, which simply informs the discount factor the environment will use to calculate the total reward.

\subsection{Observation and Action Spaces}
RRDDL and Gym have different representations of the state and action spaces. While RDDL always expects to receive the full vector of actions even if \textit{max-nondef-actions} indicates a lower number than the available number of actions, Gym interacting agents are requested to supply only the desired actions without explicitly keeping track of all the actions in the problem and their default values.
In order to mitigate these issues pyRDDLGym implements actions and states (and observations) in a \textit{Gym.Spaces.Dict} object where the keys are the action/state/observation grounded name, and the value is the intended value to pass to the simulation in case of actions, or the state/observation value in the current time step. When an agent is requested to pass actions to the environment, it is required to pass only the desired action, and not the full list of actions, the environment will augment the agent's specified action with the remainder of the actions with their default values before evaluation of the CPFs.

The conversion of RDDL types to Gym spaces is intuitive. Real valued fluents are represented as \textit{Gym.Spaces.Box}, integers are converted to \textit{Gym.Spaces.Discrete}, and Booleans are converted to \textit{Gym.Spaces.Discrete(2)}. The bounds are according to the constraints specified in the \textit{action-preconditions} block, or \textit{Numpy.inf} to denote that there is no bounding value. The \textit{action\_space} and \textit{observation\_space} will return the actions and observations respectively, where the dictionary key is the action/state/observation grounded name, and the value is the appropriate \textit{Gym.Spaces}. Sampling from the environment \textit{action\_space} will always return a valid value as the type is a two-way transformation between Gym.Spaces and RDDL types, and the bounds informed by the environment property are also in accordance with the \textit{action-preconditions} specification. In any case, the CPFs also serve as a last line of defense against values that are out of bounds.

\subsubsection{From Lifted to Grounded}
RDDL is a lifted description of a problem, i.e., the domain block defines operator schemes with parameters in contrast to explicit variables. Thus the expression 
\begin{equation}
    fluent\_cpf'(?type) = expr;
\end{equation}
is a template for all objects of type $?type$. The objects are a concrete realization of a problem and so defined in the instance. E.g., for $o_1,\ o_2$ of type $type$ the grounded realization for the objects will be
\begin{equation}
\begin{aligned}
    &fluent\_cpf'(o_1) = expr; \\
    &fluent\_cpf'(o_2) = expr;
\end{aligned}
\end{equation}
For both state/observation and action fluents, a name conversion takes place when grounding a lifted fluent with arguments, to a grounded fluent. The conversion of lifted fluent names to grounded fluent names is done with underscores. As underscores are a valid character for naming in RDDL, fluents are separated from parameters by triple underscore (\_\_\_), and parameters are separated from each other by double underscore (\_\_).
I.e., a lifted fluent with two arguments will be grounded as follows
\begin{equation}
    \begin{aligned}
        fluent(type_1, type_2)
                    \rightarrow fluent(o_1, o_2) \rightarrow fluent\_\_\_o_1\_\_o_2,
    \end{aligned}
\end{equation}
where $o_1$ and $o_2$ are objects of types $type_1$ and $type_2$ respectively. $fluent(type_1, type_2)$ is the lifted template, $fluent(o_1, o_2)$ is the grounded variable, and $fluent\_\_\_o_1\_\_o_2$ is the exposed variable name in pyRDDLGym.

\subsubsection{MDPs and POMDPs}
pyRDDLGym supports both MDPs and POMDPs. If no observation fluent is declared in the RDDL files, the framework will return the full state. In the case where observation fluents are present in the RDDL, only the observation fluents are returned by the framework. An observation fluent should be defined for a state fluent if it is observed; it is not possible to flag a state fluent as observed. Although this might seem like an unnecessary duplication of fluents, it allows for increased flexibility in specifying observations, e.g., deterministic observation, stochastic observation, etc. Due to the conversion of defining observations over the transitioned state, there is no observation at time zero, only the initial state in the case of MDPs.

\subsection{Single Time Step Evolution}
In a standard MDP, variable values are computed at discrete time steps, leading to the evolution of the process through repeated execution of the single-step transition function. Consequently, the \textit{step()} method in pyRDDLGym evaluates the CPFs defined in the RDDL problem description for all fluents in the instance.
The \textit{step()} method in pyRDDLGym undergoes four main steps. It first verifies if the actions are within the allowed range, raising a warning or exception accordingly. Then, it sorts the CPFs based on their level determined by dependency analysis. Next, it evaluates the CPF expressions using the current states and actions. Finally, it checks the state against the state invariants to ensure a legal transition and identify terminal states. If any state invariants are violated, indicating a design error, an exception is raised, and the episode is terminated, indicating it should not be used for learning.

\subsection{Resetting the Environment}
When \textit{reset()} is called the simulation is simply reverted to the initial state as specified in the instance block of the RDDL. Note that there is no randomness in the \textit{reset()} method, if one desires to reset to a different initial state, a new environment must be instantiated with a new instance. Naturally, only the initial state will be identical between episodes of the same environment, the rest of the steps will be stochastic as per the dynamics specified in that RDDL domain's CPFs.
In the case of MDP the \textit{reset()} method will return the initial values of the states, and in the case of POMDP, a dictionary with the observations will be returned, where all the values are set to the Python value \textit{None}.

\subsection{Visualization}
\begin{figure} [t]
\centering
\includegraphics[width=1.0\textwidth]{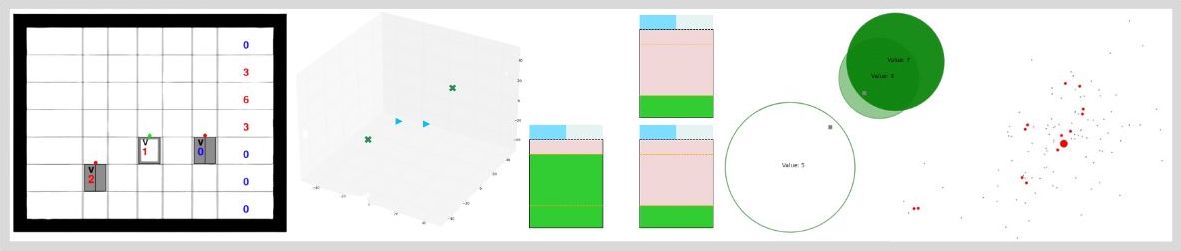}
\caption{Examples of environments implemented in pyRDDLGym. From left to right: elevators, UAV, power unit commitment, Mars rover and recommender systems.}
\label{fig:viz}
\end{figure}

pyRDDLGym also supports visualization through the Gym's \textit{render()} method. By invoking the \textit{render()} method, a visualization of the current state is displayed on the screen, and an image object is returned to the user. The default built-in visualizer that is implemented in pyRDDLGym for every environment is called \textit{TextViz}, which generates an image with a textual description of the observations and their current values. In order to create a user-defined visualizer, all is required is to implement the \textit{pyRDDLGym.Visualizer.StateViz} interface, and specify the visualization object to the environment with the method \textit{set\_visualizer($<$VizObject$>$)}. Some samples of the built-in visualizations provided with pyRDDLGym are shown in figure \ref{fig:viz}.

\section{pyRDDLGym Beyond the Engine}

\subsection{Built-in Environments}
pyRDDLGym offers advanced visualizations for all example environments. Except for the Fire Fighting domain, which is a classical RDDL domain used in the pyRDDLGym tutorial, all other examples involve continuous or hybrid spaces. All the domains have well-defined internal structures (factored spaces) that are explicitly specified in the RDDL files. The initial goal of the framework was to promote hybrid methods that combine reinforcement learning and model-based planning. Thus, we believe it's time to establish a benchmark for problems that require more than just model-free learning or combinatorial search. As pyRDDLGym is a versatile framework, all RDDL domains from previous IPCs can be easily imported by providing the relevant domain and instance files \cite{Sanner:RDDLsim}. It's worth noting that all the past IPC problems can be accessed and imported into pyRDDLGym through the "rddlrepository" repository, which can be found at: \url{https://github.com/pyrddlgym-project/rddlrepository}.

\renewcommand{\arraystretch}{1.5}
\begin{center}
\begin{table*}[h!]
\centering
\begin{tabular}{P{5.8cm} P{2cm} P{2cm} P{5cm}}
 \ChangeRT{1mm}
 \textbf{Domain Name} & \textbf{Action Space} & \textbf{State Space} & \textbf{Source} \\ [0.5ex] 
 \ChangeRT{1mm}
 Cart-pole Continuous & Continuous & Continuous & \citet{barto1983neuronlike} \\ 
 \hline
 Cart-pole Discrete & Discrete & Continuous & \citet{barto1983neuronlike} \\
 \hline
 Elevators & Discrete & Discrete & Ours \\
 \hline
 Mars Rover & Mixed & Mixed & \citet{taitler2019combined} \\
 \hline
 Mountain-car & Continuous & Continuous & \citet{Moore90efficientmemory-based} \\ [1ex] 
 \hline
 Power Unit Commitment Discrete & Discrete & Mixed & \citet{abdou2018unit} \\ [1ex] 
 \hline
 Power Unit Commitment Continuous & Continuous & Continuous & \citet{abdou2018unit} \\ [1ex] 
 \hline
 Racing Car & Continuous & Continuous & Ours \\ [1ex] 
 \hline
 Recommender Systems & Discrete & Continuous & \citet{pmlr-v119-mladenov20a} \\ [1ex] 
 \hline
 Fire Fighting & Discrete & Discrete & \citet{karafyllidis1997model} \\ [1ex] 
 \hline
 UAV Continuous & Continuous & Continuous & Ours \\ [1ex] 
 \hline
 UAV Discrete & Discrete & Continuous & Ours \\ [1ex] 
 \hline
 UAV Mixed & Mixed & Continuous & Ours \\ [1ex] 
 \hline
 Supply Chain & Discrete & Discrete & \citet{kemmer2018reinforcement} \\ [1ex] 
 \hline
 Traffic & Discrete & Continuous & \citet{lin2009simplified} \\ [1ex]
 \ChangeRT{1mm}
\end{tabular}
\vspace{3mm}
\caption{List of domains currently included in pyRDDLGym. For each environment, we report the original source behind the RDDL files and the type of action and state spaces,
whether they are fully discrete, continuous, or mixed discrete-continuous.}
    \label{tab:examples}
\end{table*}
\end{center}

Currently, there are $15$ environments implemented in pyRDDLGym. Adapted from the previous simulator \cite{Sanner:RDDLsim}, 
the Fire Fighting domain is a discrete domain. 
It was added to pyRDDLGym as an introduction domain for learning pyRDDLGym and RDDL. Mars Rover domain is a dynamic version of the standard MAPF problem \cite{standley2010finding},
with dynamic properties and inspiration from \cite{taitler2019combined} and \cite{fernandez2018scottyactivity}. 
The Power Unit Commitment example has two versions, a discrete action version which was taken as it is from the previous 2014 IPC, and a continuous action version.
There are three original domains; elevators, which is a discrete domain, Racing car is a continuous control domain, in which completely different racing tracks can be constructed by playing with the non-fluents. The UAV domain is the third, and it has three versions as well. The three versions differ by the type of action space. The first is fully continuous, the second is fully discrete, and the third has a mixed discrete-continuous action space. The dynamics in the UAV are a slightly simplified version of the full model described in \citet{hull2007fundamentals}.
Two domains were adapted from OpenAI Gym classic control domains, in order to have some familiar domains and to show how simple the conversion process into RDDL is. Mountain Car, and Cart-pole. Cart-pole comes in two versions: one for continuous actions and one for discrete actions.
Another domain is a Recommander System \cite{pmlr-v119-mladenov20a}, which is not a classical control or Operation Research problem but has gained focus in the last years, and in real life scales up to millions of objects, which by itself is a challenge to decision-making algorithms.
The last domain is a traffic domain, modeled after the QTM/BLX models \cite{guilliard2016nonhomogeneous, lin2009simplified}. It is a macroscopic flow and traffic model that can be scaled from a single isolated intersection to a large network of signalized intersections.
The list of environments with the details is presented in Table \ref{tab:examples}. Additionally, example of how to go from problem definition to a working pyRDDLGym environment is given in appendix \ref{sec:appndx}.

\subsection{The Example Manager}
In order to access the built-in examples in pyRDDLGym, the unified interface of supplying domain and instance files, and a visualizer object is used. The ExampleManager object is where the domains, instances, and visualizers for all examples are registered. The ExampleManager is documented in the pyRDDLGym documentation, but it has several key methods.
The first is a \textit{ListExamples()}, a static method that lists all the examples with a short description. The names of the domains as listed in that method can be used to instantiate the ExampleManager object and access the example details. Then the methods \textit{get\_domain()},
\textit{get\_instance(\#)} can be used to get the path to the domain and instance files. Finally \textit{get\_visualizer()} returns the dedicated visualization object for the example.

\begin{figure}[t!]
     \centering
     \begin{subfigure}[b]{0.49\textwidth}
         \centering
         \includegraphics[width=\textwidth]{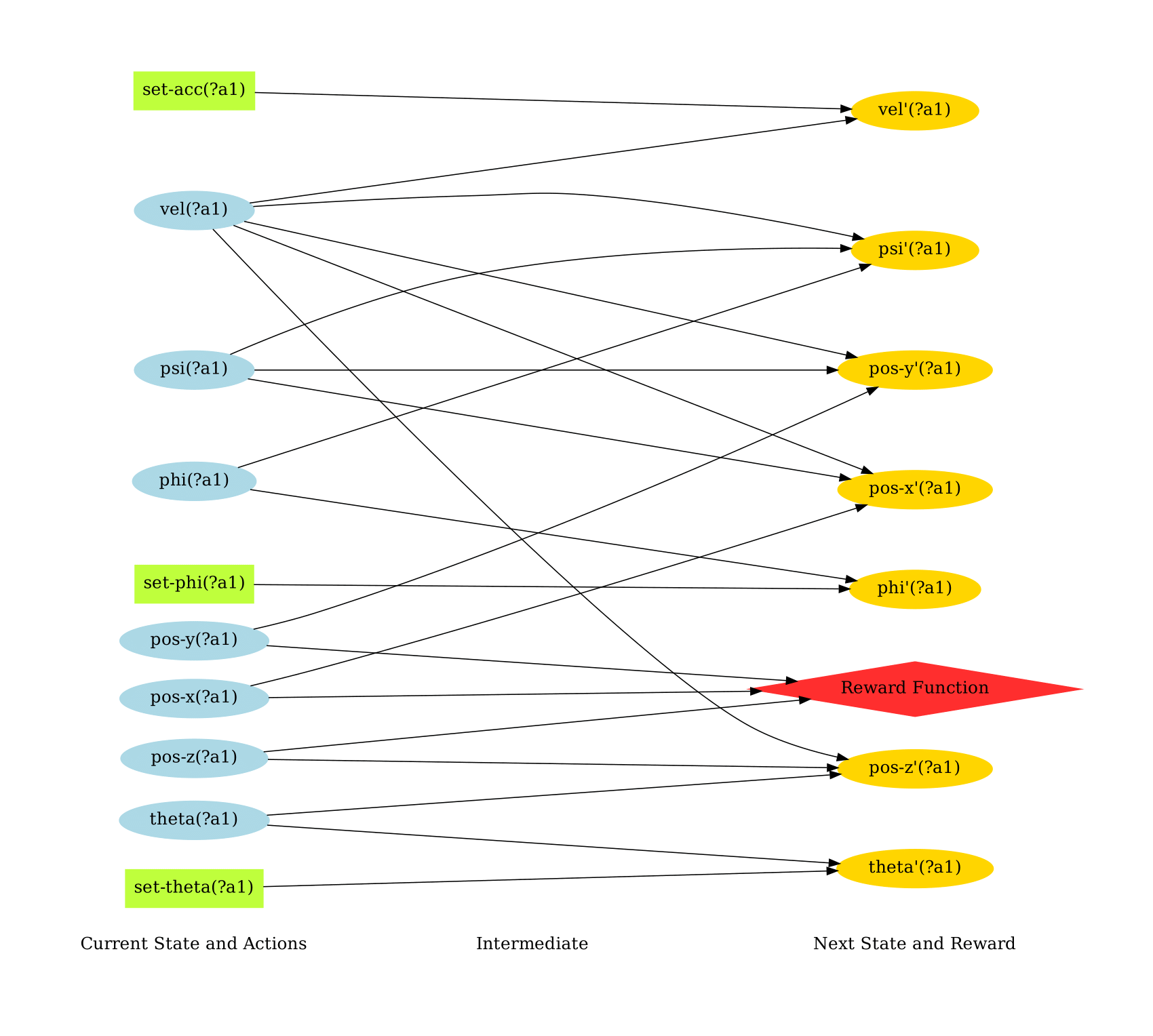}
         \caption{DBN generated for the UAV domain}
         \label{fig:dbn}
     \end{subfigure}
     \begin{subfigure}[b]{0.49\textwidth}
         \centering
         \includegraphics[width=\textwidth]{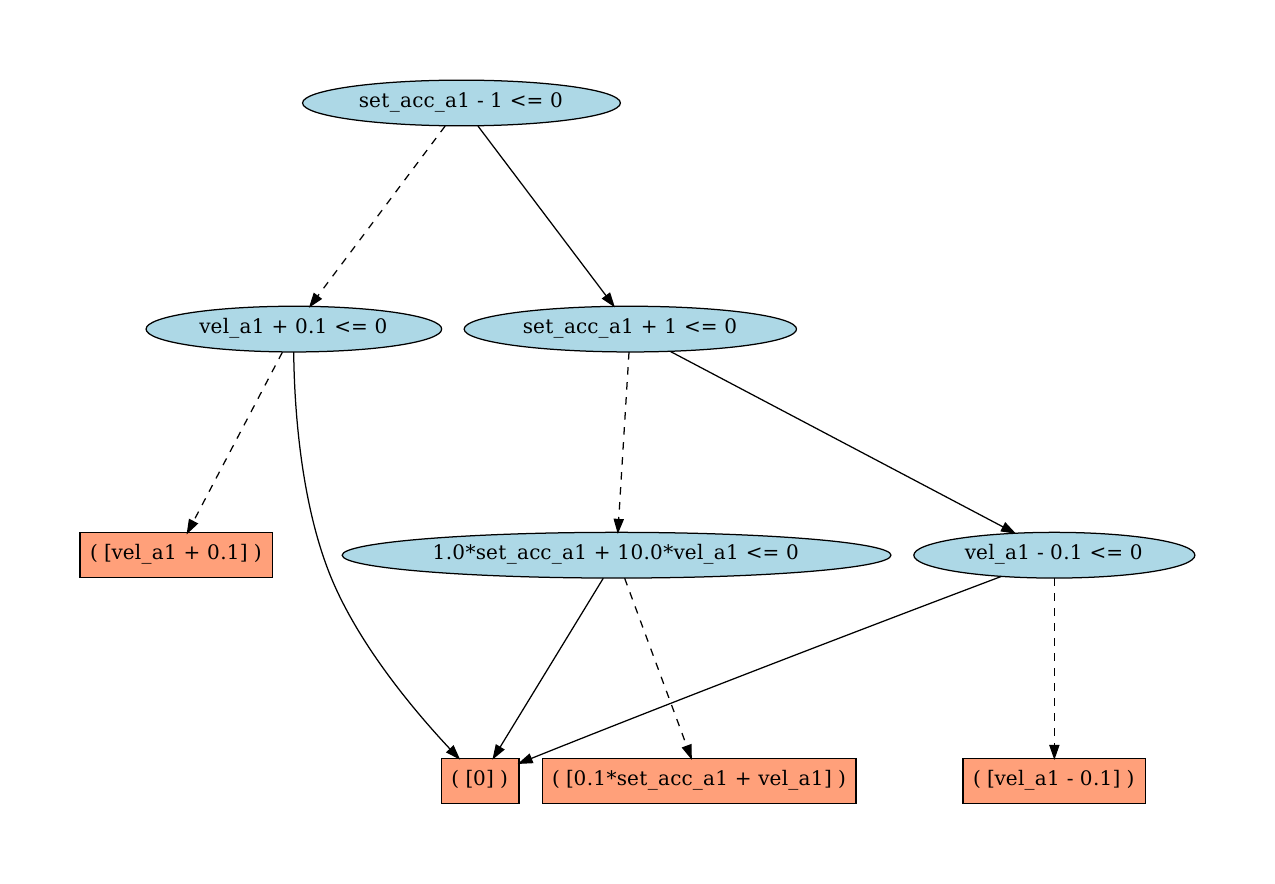}
         \caption{xADD generated for the velocity state in the UAV domain}
         \label{fig:xadd}
     \end{subfigure}
        \caption{pyRDDLGym auxiliary tools}
        \label{fig:AUX}
\end{figure}

\subsection{Auxiliary Tools}
\subsubsection{XADDs}
XADD (eXtended Algebraic Decision Diagram) \cite{sanner2011symbolic} enables compact representation and operations with symbolic variables and functions. In fact, this data structure can be used to represent CPFs defined in a RDDL domain once it is grounded for a specific RDDL instance. pyRDDLGym can generate XADDs for all states/observations in the grounded domain for the use of planning agents.

\subsubsection{Dynamic Bayes Nets Visualization}
With the XADD compilation of a given domain instance, one can easily visualize the dependencies between different fluents. For this purpose, a Dynamic Bayes Nets (DBNs) visualization tool is also provided in the framework. This tool provides a way to produce diagrams similar to an influence diagram.


\subsection{Model-Based Planner}
While NumPy \cite{harris2020array} serves as the default backend for pyRDDLGym, it also offers support for JAX \cite{jax2018github} as an alternative that can handle gradients. This capability enables planning using back-propagation approaches \cite{wu2017scalable, bueno2019deep}. Moreover, pyRDDLGym features an implemented JAXPlanner, which encompasses the entire range of the RDDL language, facilitates GPU execution, and addresses stochastic problems. In practical terms, given model, a roll-out described by an influence graph can be generated, and the total reward or other performance metrics, such as risk \cite{patton2022distributional}, can be calculated and differentiated with respect to the inputs.

One notable feature of JAXPlanner is its ability to handle hybrid continuous-discrete state and action spaces. To facilitate derivatives of CPFs with respect to action-fluents, JAXPlanner replaces functional dependencies $F_i$ with differentiable relaxations, formalized as families of functions $\lbrace f_{i,\tau}: \tau > 0\rbrace$ indexed by some hyper-parameter $\tau$. The variables $\tilde{X}_i = f_{i,\tau}(\mathrm{Pa}(\tilde{X}_i))$ with $\tilde{X}_1 = X_1$ then define an equivalent DAG $\tilde{\mathcal{G}} = (\tilde{X}, \mathrm{Pa})$ with the same edges as $\mathcal{G}$ but where nodes $X_i$ are replaced by $\tilde{X}_i$. Thus, a differentiable model approximation is achieved.

Given that Boolean logic in RDDL lacks inherent differentiability, it becomes essential to discover functions $f_{i,\tau}$ that effectively approximate Boolean logic. One approach is to substitute Boolean operations with t-norms \cite{hajek2013metamathematics}. JAXPlanner incorporates t-norm approximations, but it also provides support for operation overloading, allowing users to define their own operators. This flexibility enables users to tailor the implementation according to their specific requirements. Specifically, a t-norm is a function $T: [0, 1]^2 \to [0, 1]$ which satisfy $4$ properties: commutativity, monotonicity, associativity, and inclusion of the identity element.
For Boolean-valued quantities $a, b$, JAXPlanner uses the following approximations:
\begin{itemize}
    \item $a \land b \approx T(a,b)$
    \item $\neg a \approx 1 - a$,
\end{itemize}
from which the other RDDL operations can be derived, e.g.:
\begin{itemize}
    \item $a \lor b \equiv \neg (\neg a \land \neg b) \approx 1 - T(1 - a, 1 - b)$
    \item $a \implies b \equiv \neg a \lor b \approx 1 - T(a, 1 - b)$
    \item $\forall\lbrace x_1, x_2, \dots x_m \rbrace \equiv x_1 \land x_2 \dots \land x_m  \approx T(x_1, T(x_2, T(\dots)))$
    \item $\exists\lbrace x_1, x_2, \dots x_m \rbrace \equiv \neg\forall\lbrace \neg x_1, \neg x_2, \dots \neg x_m \rbrace \approx 1 - T(1 - x_1, T(1 - x_2, T(\dots)))$.    
\end{itemize}
The conditional branching statement such as
\begin{equation*}
    \textrm{if } (c) \textrm{ then } a \textrm{ else } b
\end{equation*}
is rewritten in our framework as
\begin{equation*}
    f(a, b, c) = c * a + (1 - c) * b,
\end{equation*}
which is a continuous function.
A popular choice for approximating relational operations such as $a > b$, $a < b$ and $a = b$ as suggested in prior literature is to use sigmoid functions. JAXPlanner opt to use the logistic sigmoid:
\begin{align*}
    &a > b \approx f_>(a, b, \tau) = \mathrm{sigmoid}((a - b) / \tau) \\
    &a = b \approx f_=(a, b, \tau) = \mathrm{sech}^2((b - a) / \tau),
\end{align*}
where $\tau$ refers to the temperature parameter.

\begin{figure*}[t]
     \centering
     \begin{subfigure}{0.30\textwidth}
         \centering
         \includegraphics[width=\textwidth]{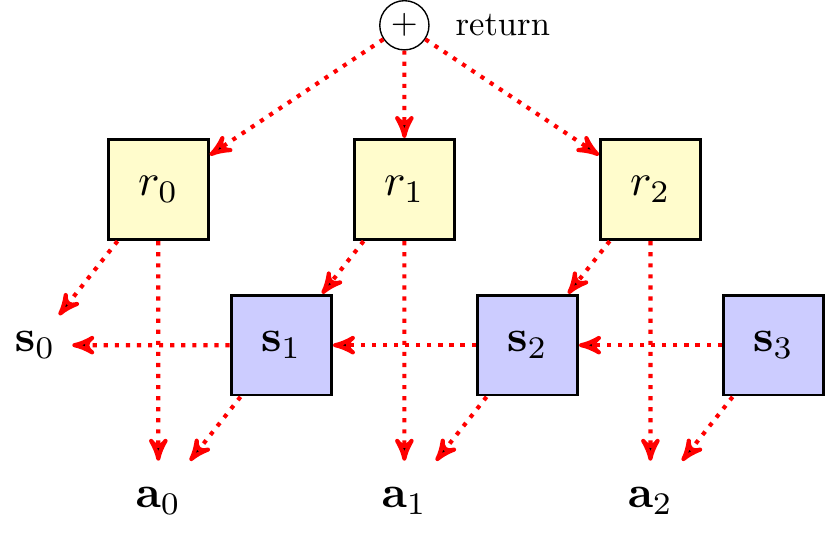}
         \caption{Straight line roll-out with gradients.}
         \label{fig:jax2}
     \end{subfigure}%
     \begin{subfigure}{0.45\textwidth}
         \centering
         \includegraphics[width=\textwidth]{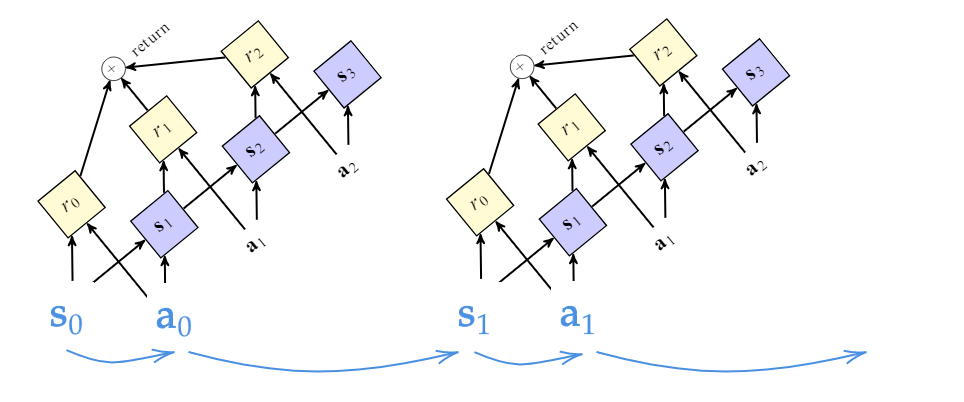}
         \caption{Re-planning over fixed time rolling horizon.}
         \label{fig:jax3}
     \end{subfigure}%
          \begin{subfigure}{0.25\textwidth}
         \centering
         \includegraphics[width=\textwidth]{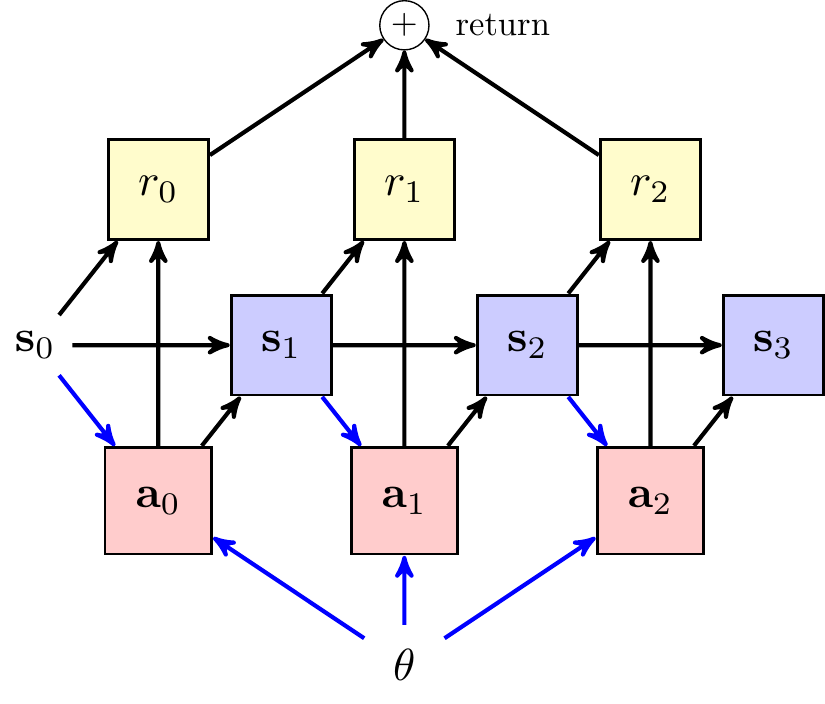}
         \caption{Deep Reactive Policies.}
         \label{fig:jax4}
     \end{subfigure}
        \caption{JAXPlanner modes of operations.}
        \label{fig:JAX}
\end{figure*}

JAXPlanner offers two operational modes, depicted in Figure \ref{fig:JAX}. The first mode involves generating a straight line plan (SLP) by performing a complete horizon rollout of the problem. It is worth noting that utilizing this method with a fixed horizon at each step produces a solution resembling model predictive control. The second mode involves employing a neural network to train a deep reactive policy (DRP), which is similar to the approach taken in \citet{bueno2019deep}.

\section{Conclusion and Future Work}

We have presented pyRDDLGym, an open-source Python framework that automatically creates OpenAI Gym environments from the RDDL domain and instance files. We hope that the availability of such a framework will help foster collaboration between researchers in the learning and planning communities. We also believe that by separating the process of problem design and the no longer needed programming task, a verified benchmark for RL and planning can be established independently of a specific platform. Finally from our own experience the generation of a brand-new environment has been accelerated significantly and the logic can be formally verified, which so far has not been possible.

\bibliographystyle{unsrtnat}  
\bibliography{references}  

\newpage
\appendix 

\section{From Math to RDDL Domain Case Study} \label{sec:appndx}
\begin{figure}[h!]
\centering
\includegraphics[width=0.46\textwidth]{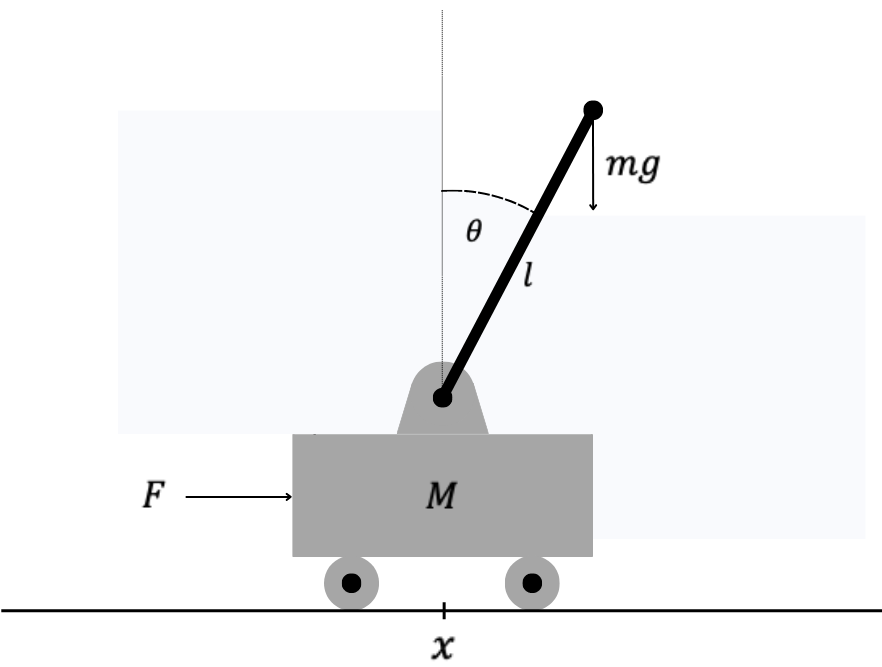}
\caption{The cart-pole balancing model}
\label{fig:cartpole}
\end{figure}

We take as a case study the classical Gym environment Cart-pole, the physical realization of the system is given in figure \ref{fig:cartpole}.

The horizontal position of the cart is represented as $x$, while the angular position of the pole is denoted as $\theta$. The velocities or rates of change are appropriately indicated by the dot notation, $\dot{x}$ and $\dot{\theta}$. The force propelling the cart along the horizontal axis is denoted as $F$, and all of these variables are time-dependent functions. The system parameters include the cart's mass ($M$), the pole's mass ($m$), the length of the pole ($l$), and the gravitational constant ($g$). Additional constants that can be incorporated are the allowable bounds for the pole angle, track length, and force limits. The Gym environment implements in python directly the system dynamics, thus, all the parameters are hard-coded within the \textit{step()} function.
To implement the same environment with the same behaviour in RDDL we should start with the dynamic equations. The state vector naturally is $[x,v_x, \theta, v_\theta]^T$. In continuous time the equations are given by \cite{barto1983neuronlike}:

\begin{equation} \label{eq:cartpole}
    \begin{aligned}
        &\dot v_\theta = \frac{g \sin \theta + \cos \theta \big(\frac{-F -ml v_\theta^2 \sin \theta}{m+M}\big)}{ l \big(  \frac{4}{3} - \frac{m\cos^2\theta}{m+M}  \big) } \\
        & \dot v_x = \frac{F + ml(v_\theta^2\sin \theta - \dot v_\theta \cos \theta)}{m+M}
    \end{aligned}
\end{equation}

Using a first order discrete approximation with sampling time $T$, and writing \eqref{eq:cartpole} as a system of first order linear system yields:
\begin{equation} \label{eq:cartpole_lin}
    \begin{aligned}
        x(k+1) &= x(k) + Tv_x(k) \\
        v_x(k+1) &= v_x(k) + Tu_x(k) \\
        \theta(k+1) &= \theta(k) + Tv_\theta(k) \\
        v_\theta(k+1) &= v_\theta(k) + T u_\theta(k)
    \end{aligned}
\end{equation}
where $u_x, u_\theta$ are virtual inputs to the cart and pole motion appropriately, defined as:

\begin{equation}
\begin{aligned} \label{eq:virtual}
    &u_\theta(k) = \frac{g \sin \theta(k) - \cos \theta(k) t(k)}{l \big(\frac{4}{3} - \frac{m \cos^2 \theta(k)}{m+M} \big)} \\
    &u_x(k) = t(k) - \frac{l m u_\theta(k) \cos \theta(k)}{m + M}
\end{aligned}
\end{equation}
and $t(k)$ is an auxiliary definition to keep the math a little cleaner:
\begin{equation} \label{eq:temp}
    t(k) = \frac{F(k) + l m  v_\theta^2(k) \sin \theta(k)}{m+M}
\end{equation}

Now, we are prepared to convert the mathematical representation into RDDL. We commence by declaring the variables involved in the problem. The gravity constant is a non-fluent with a default value of $9.8$. Since the Gym environment features a discrete action space, where 1 represents a constant force to the right and 0 represents a constant force to the left, we define a non-fluent that represents the force. All other changing constants, such as the masses of the cart and pole, pole length, time step, and the limits on the cart position and pole angle, are also non-fluents. The auxiliary functions described in equations \eqref{eq:virtual} and \eqref{eq:temp} are considered interm-fluents. The four variables that define the state are referred to as state-fluents, while the input force applied to the cart's side is an action-fluent. The functional definition is presented within the "cpfs" block, where the next time step's state variables are denoted with a prime symbol, e.g., $x'$ instead of $x{(k+1)}$. The "cpfs" block directly implements equations \eqref{eq:cartpole_lin}-\eqref{eq:temp}. An additional interm-fluent is defined for the purpose of translating the action index to force value, implemented in the first CPF at the "cpfs" block.

\begin{figure*}[t]
\centering
\includegraphics[width=1.0\textwidth]{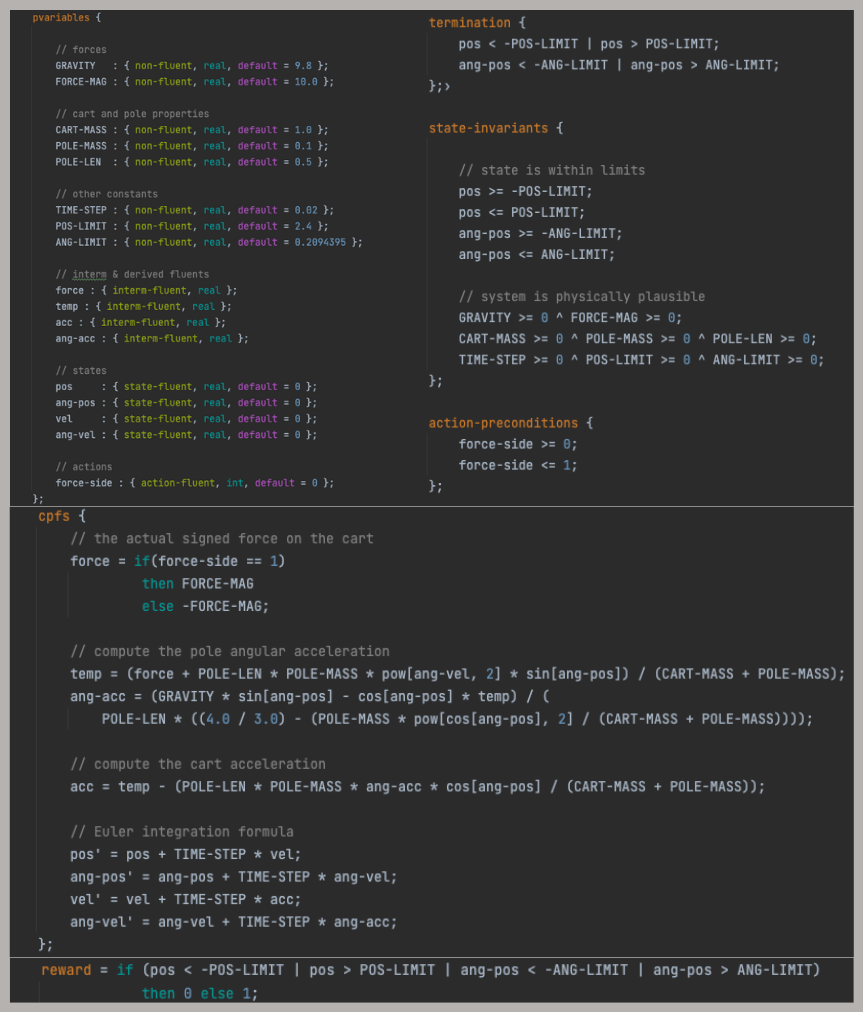}
\caption{The Cart-pole problem depicted in RDDL format, featuring variable definitions (top left), mathematical functions and state evolution (middle section), termination conditions and constraints (top right), and reward function definition (bottom).}
\label{fig:cartpole_rddl}
\end{figure*}

The simulation terminates when the "termination" block evaluates to "true," indicating that either the cart has moved outside the track or the pole angle has exceeded the angle limit. The state-invariants serve as safeguards to ensure that no instance places the cart or pole outside the feasible ranges at the initial state, and that the problem's constants are non-negative. It is permissible to define an instance on the Moon, for example, with a gravity coefficient of $1.64 [m/s^2]$. The action-preconditions ensure that the only acceptable actions are the integers $1$ and $2$. Lastly, the reward function grants the agent a positive value of $1.0$ for every time step in which the cart remains on the track and the pole does not fall outside the angle limits. Each aspect of the domain is defined mathematically, without the need to implement complex functionality or termination decisions. Typically, an instance file would solely set the initial state of the Cart-pole system, although one has the option to adjust the constants (input force, cart mass, pole length, etc.) according to their requirements. The complete domain description can be found in Figure \ref{fig:cartpole_rddl}.

\end{document}